\documentclass{article}
\usepackage{float}
\usepackage{amsmath}
\usepackage{PRIMEarxiv}
\usepackage{makecell}
\usepackage[utf8]{inputenc} 
\usepackage[T1]{fontenc}    
\usepackage{hyperref}       
\usepackage{url}            
\usepackage{booktabs}       
\usepackage{amsfonts}       
\usepackage{nicefrac}       
\usepackage{microtype}      
\usepackage{lipsum}
\usepackage{fancyhdr}       
\usepackage{graphicx}       
\graphicspath{{media/}} 
\usepackage{xcolor}
\usepackage{authblk}
\usepackage{hyperref}
\usepackage{multirow} 
\usepackage{longtable} 
\usepackage{tabularx} 
\usepackage{booktabs}  
\begin{document}

\title{Can Large Language Models Function as Qualified Pediatricians? A Systematic Evaluation in Real-World Clinical Contexts}
\author[1,†]{Siyu Zhu}
\author[2,†]{Mouxiao Bian}
\author[1,†]{Yue Xie}
\author[3]{Yongyu Tang}
\author[2]{Zhikang Yu}
\author[2]{Tianbin Li}
\author[2,4]{Pengcheng Chen}
\author[2]{Bing Han}
\author[2,*]{Jie Xu}
\author[1,*]{Xiaoyan Dong}
\affil[1]{\textit{
Shanghai Children's Hospital, School of Medicine,Shanghai Jiao Tong University \\
  Shanghai, China
}}
\affil[2]{\textit{
    Shanghai Artificial Intelligence Laboratory\\
    Shanghai, China
}}
\affil[3]{\textit{
   Longhua Hospital Shanghai University of Traditional Chinese Medicine\\
    Shanghai, China
}}
\affil[4]{\textit{
   University of Washington\\
    Washington, USA
}}
\footnotetext[1]{†These authors contributed equally.}
\footnotetext[2]{*Correspondence: 
Xiaoyan Dong(dongxy@shchildren.com.cn),
Jie Xu (xujie@pjlab.org.cn)
}

\maketitle
\begin{abstract}
With the rapid rise of large language models (LLMs) in medicine, a key question is whether they can function as competent pediatricians in real-world clinical settings. We developed PEDIASBench (Pediatric Evaluation of Dynamic Intelligence, Adaptability, and Safety Benchmark), a systematic evaluation framework centered on a knowledge-system framework and tailored to realistic clinical environments. PEDIASBench assesses LLMs across three dimensions: application of basic knowledge foundational knowledge, dynamic diagnosis and treatment capability, and pediatric medical safety and medical ethics. We evaluated 12 representative models released over the past two years, including GPT-4o, Qwen3-235B-A22B, and DeepSeek-V3, covering 19 pediatric subspecialties and 211 prototypical diseases.State-of-the-art models performed well on foundational knowledge, with Qwen3-235B-A22B achieving over 90\% accuracy on licensing-level questions, but performance declined \~15\% as task complexity increased, revealing limitations in complex reasoning. Multiple-choice assessments highlighted weaknesses in integrative reasoning and knowledge recall. In dynamic diagnosis and treatment scenarios, DeepSeek-R1 scored highest in case reasoning (mean 0.58), yet most models struggled to adapt to real-time patient changes. On pediatric medical ethics and safety tasks, Qwen2.5-72B performed best (accuracy 92.05\%), though humanistic sensitivity remained limited.
These findings indicate that pediatric LLMs are constrained by limited dynamic decision-making and underdeveloped humanistic care. Future development should focus on multimodal integration (text, imaging, physiological signals) and a clinical feedback–model iteration loop to enhance safety, interpretability, and human–AI collaboration. While current LLMs cannot independently perform pediatric care, they hold promise for decision support, medical education, and patient communication, laying the groundwork for a safe, trustworthy, and collaborative intelligent pediatric healthcare system.

\end{abstract}

\section{Introduction}
With the advent of the Transformer architecture and the release of the GPT series, LLMs have demonstrated human-like reasoning abilities and achieved impressive performance in medical examinations. For example, GPT-4 has achieved passing scores on simulated United States Medical Licensing Examination (USMLE) tasks and can provide detailed explanations for its answers\cite{ferber2025development}. These cross-domain capabilities have drawn significant attention in the medical community, with applications spanning documentation summarization, information retrieval, and clinical decision support\cite{riedemann2024path}\cite{singhal2025toward}.

As LLMs become increasingly integrated into healthcare, their potential to enhance pediatric diagnostics and decision-making is attracting attention. Pediatric medicine differs from adult medicine in scope and complexity, encompassing rapidly changing developmental stages, disease heterogeneity, and communication challenges unique to children and families. Recent studies show that LLMs can help reduce pediatric medication dosing errors\cite{levin2025can}, support individualized treatment planning by integrating updated clinical guidelines and case data\cite{li2024integrated}\cite{huang2024assessment}, and improve efficiency in clinical documentation through natural language processing. Conversational AI tools can facilitate parent–child–clinician communication by translating complex medical concepts into accessible language, and may even supplement pediatric mental health education through bias-free dialogue environments\cite{fahrner2025generative}\cite{barile2024diagnostic}\cite{mansoor2025conversational}.
In medical education, LLMs are emerging as valuable learning tools. GPT-4, for instance, can already achieve or surpass average human performance in several physician licensing examinations\cite{katz2024gpt}\cite{kipp2024gpt}\cite{liu2024performance}. Nevertheless, their reliability in specialized pediatric assessments remains inconsistent. Beam et al. reported that ChatGPT-3.5 correctly answered only 46\% of 936 neonatal medicine questions\cite{beam2023performance}, indicating substantial limitations. Furthermore, clinical trials reveal that LLMs, though individually accurate, do not yet translate into measurable improvements when used as diagnostic assistants by physicians\cite{goh2024large}.

Pediatrics, however, presents unique challenges. It encompasses an exceptionally broad disease spectrum\cite{chen2024matching}, involves patients in continuous developmental stages\cite{black2017early}, and requires family-centered communication approaches\cite{seniwati2023effects}\cite{hodgson2024child}\cite{mccarthy2022family}. Pediatric practice is characterized by precise weight-based dosing\cite{arnold2009personalized}\cite{kearns2003developmental} and age-adapted diagnostic communication\cite{stein2019communication}Pediatric medicine differs from adult medicine in scope and complexity, encompassing rapidly changing developmental stages, disease heterogeneity, and communication challenges unique to children and families. Recent studies show that LLMs can help reduce pediatric medication dosing errors\cite{klassen2008children}\cite{chng2025ethical}.

Existing medical benchmarks, such as Pediabench\cite{zhang2024pediabench} and general-purpose datasets including MedQA\cite{jin2021disease}, and PubMedQA\cite{jin2019pubmedqa}, provide partial insights but lack systematic evaluation of pediatric-specific clinical reasoning. Pediatric medicine differs from adult medicine in scope and complexity, encompassing rapidly changing developmental stages, disease heterogeneity, and communication challenges unique to children and families. Recent studies show that LLMs can help reduce pediatric medication dosing errors\cite{wang2024large}.

Given these challenges, we proposed PEDIASBench, a systematic, clinically authentic, and dynamically adaptive evaluation framework that integrates a three-dimensional pediatric competency system(Figure 1). This system mirrors the real-world capabilities required of pediatricians, encompassing application of basic medical knowledge, dynamic diagnostic and treatment capacity, and pediatric medical safety and ethics.The first two dimensions reflect professional skills—rooted in clinical reasoning and procedural competence—while the latter forms the ethical foundation of safe and humanistic pediatric care.

To ensure comprehensive coverage across pediatric subspecialties, PEDIASBench incorporates 19 departments, including pediatric internal medicine and pediatric surgery, encompassing 211 representative diseases. Each disease module integrates authentic clinical cases and corresponding question types drawn from four standardized examination, thus bridging standardized medical education assessments with real clinical complexity\cite{frank2010competency}\cite{ten2010medical}. This structure allows the benchmark to evaluate not only factual recall but also adaptive reasoning in uncertain, context-rich scenarios—an essential characteristic of clinical expertise\cite{norman2005research}.

Beyond professional competence, PEDIASBench embeds ethical and communicative dimensions, addressing clinical ethics, patient safety, and doctor–patient communication. These components align with contemporary pediatric ethics frameworks, emphasizing beneficence, nonmaleficence, autonomy, and justice in child healthcare. By integrating these dimensions, PEDIASBench extends beyond knowledge-based testing to offer a holistic assessment of a model’s readiness to function as a trustworthy, safe, and empathetic collaborator in pediatric practice\cite{li2023trustworthy}. 

In summary, PEDIASBench establishes a comprehensive, multidimensional, and ethically grounded paradigm for evaluating large language models within the pediatric clinical context. By integrating foundational medical knowledge, dynamic diagnostic reasoning, and ethical–safety dimensions into a unified assessment framework, it reflects the authentic cognitive and moral demands placed upon pediatricians in real-world practice\cite{frank2010competency}. Unlike conventional benchmarks that isolate factual recall or task-specific accuracy, PEDIASBench advances toward a competency-based, context-aware evaluation model aligned with modern medical education principles. This integration not only facilitates cross-specialty performance analysis across 19 pediatric disciplines but also underscores the indispensable role of empathy, communication, and safety in the responsible deployment of medical AI systems. As such, PEDIASBench provides a foundational pathway toward verifying whether large language models can truly approach the standards of a qualified pediatric practitioner—one who combines precision with compassion, and intelligence with integrity.
\begin{figure}
    \centering
    \includegraphics[width=1\linewidth]{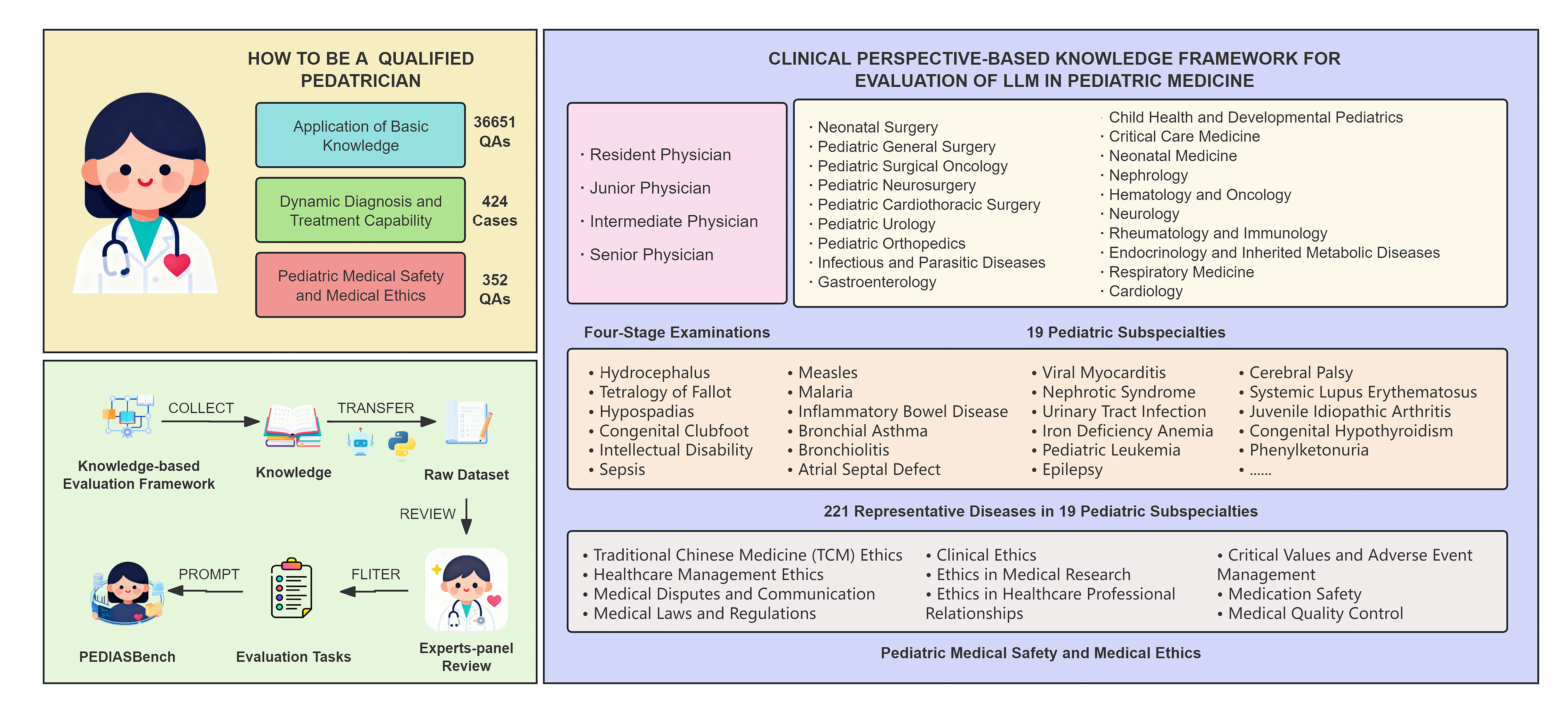}
    \caption{Pediatric Evaluation of Dynamic Intelligence, Adaptability, and Safety Benchmark}
    \label{fig:introduction}
\end{figure}
\section{Evaluation Models and Framework}
\subsection{Evaluation Models}
A total of 12 representative LLMs released within the past two years were selected, encompassing both commercial and open-source models, as well as small-scale (<10B parameters) and large-scale (>200B parameters) architectures. The detailed information is presented in Table \ref{tab:overview_of_models}.

\begin{table}[ht]
    \centering
    \caption{Overview of Evaluated Models}
    \label{tab:overview_of_models}
    \begin{tabularx}{\textwidth}{@{}p{2.5cm} p{5cm} p{2.5cm} p{2.5cm} p{2cm}@{}} 
    \toprule
    \textbf{Name} & \textbf{Model} & \textbf{Size} & \textbf{Release Date} & \textbf{Model Type} \\  
    \midrule
    \multicolumn{5}{c}{\textbf{Open Source}}\\ 
    DeepSeek & DeepSeek-R1 & 671B & 2025-05 & Text Only \\  
    \addlinespace
    DeepSeek & DeepSeek-V3 & 671B & 2024-12 & Text Only \\
    \addlinespace
    GLM & GLM-4-9B-Chat & 9B & 2024-11 & Text Only \\
    \addlinespace
    InternLM & Intern2.5-7B-Chat & 7B & 2024-06 & Text Only \\
    \addlinespace
    Llama & Llama-4-Maverick & \~400B & 2025-04 & Multimodal \\
    \addlinespace
    Mistral AI & Mistral-Small-3.1-24B-Instruct & 24B & 2025-03 & Text Only \\
    \addlinespace
    Qianwen & Qwen2.5-72B & 72B & 2024-09 & Text Only \\
    \addlinespace
    Qianwen & Qwen3-32B & 32B & 2025-04 & Text Only \\
    \addlinespace
    Qianwen & Qwen3-235B-A22B & 235B & 2025-04 & Text Only \\
    \addlinespace
    \multicolumn{5}{c}{\textbf{Closed Source}}\\
    Gemini & Gemini-2.5-Flash & — & 2025-04 & Multimodal \\
    \addlinespace
    GPT & GPT-4o-2024-11-20 & \~200B & 2024-11 & Multimodal \\
    \addlinespace
    GPT & GPT-5-Mini-2025-08-07 & — & 2025-08 & Multimodal \\
    \bottomrule
    \end{tabularx}
\end{table}

\subsection{Evaluation Framework}
The framework was constructed to reflect the competency architecture of a qualified pediatrician within real-world clinical contexts. It comprises three principal domains: Application of Basic Knowledge, Dynamic Diagnostic and Therapeutic Capacity, and Medical Ethics and Patient Safety.The Medical Ethics and Patient Safety domain extends beyond ethical conduct to incorporate medical communication, empathy, professionalism, and patient-centered safety awareness, thereby integrating both humanistic and ethical dimensions of pediatric care.While the first two domains primarily evaluate professional and cognitive competencies, the third domain serves as the moral and behavioral foundation that underpins the integrity, empathy, and safety of pediatric practice.
To ensure comprehensive assessment, the first two domains were further subdivided into 19 pediatric subspecialties (covering both pediatric internal medicine and pediatric surgery). A total of 211 representative diseases were selected across these subspecialties, with corresponding clinical cases and four levels of standardized clinical examinations curated for each.
The medical safety and ethics dimension included 10 subcategories, such as clinical ethics, patient communication, informed consent, medical quality control, and pediatric safety management.
\subsection{Evaluation Metrics}
\subsubsection{Single-choice Questions}
Performance on single-choice items was quantified by accuracy, defined as(Equation \ref{Accuracy}):
\begin{equation}
\label{Accuracy}
\text{Accuracy} = \frac{\text{TP} + \text{TN}}{\text{TP} + \text{TN} + \text{FP} + \text{FN}}
\end{equation}
 where TP (True Positives) and TN (True Negatives) represent correctly predicted cases, while FP (False Positives) and FN (False Negatives) denote incorrect classifications. Accuracy thus measures overall correctness of discrete predictions.
\subsubsection{Multiple-choice Questions}
Performance was measured by the F1 score (Equation \ref{F1_score}), combining precision and recall and accuracy (Equation \ref{Accuracy} ).
\begin{equation}
\label{F1_score}
F_1 = 2 \times \frac{\text{Precision} \times \text{Recall}}{\text{Precision} + \text{Recall}}
\end{equation}

Where:\\
- Precision represents the proportion of samples predicted as positive that are actually positive(Equation \ref{precision}):
 \begin{equation} 
 \label{precision}
  \text{Precision} = \frac{TP}{TP + FP}
  \end{equation}
- Recall represents the proportion of actual positive samples that are correctly predicted as positive(Equation \ref{recall}):
\begin{equation}
 \label{recall}
  \text{Recall} = \frac{TP}{TP + FN}
  \end{equation}

In the above formulas, $TP$ stands for True Positives, $FP$ stands for False Positives, and $FN$ stands for False Negatives.

\subsubsection{Short-answer Questions}
Short-answer (open-ended) items were evaluated on two dimensions (Equation \ref{total_cal}):
\begin{equation}
 \label{total_cal}
Total = 0.7\times Total_{\text{Macro Recall}} + 0.3 \times Total_{\text{BERTScore}}
 \end{equation}

where:  
- \textbf{Answer points} refer to key information points (e.g., critical details related to diagnosis, treatment, and care) pre-labeled by clinical experts based on clinical guidelines and practical experience. These points are defined as essential elements that a complete and accurate answer should cover.  

- \textbf{Macro Recall} is calculated as the average recall rate of all answer points within each question. Specifically, for each pre-labeled answer point $i$ in a question, its recall rate is computed by:  
  \begin{equation*}
  \text{Recall}_i = \frac{\text{Number of times answer point } i \text{ is covered in the model's output}}{\text{Total number of answer point } i \text{ in the reference answer}}
  \end{equation*}  
  $Total_{\text{Macro Recall}}$ is then the arithmetic mean of $\text{Recall}_i$ across all answer points within the question, implemented via Python (using custom scripts to match and count key points).

- \textbf{BERTScore} serves as an auxiliary tool: when there are expression differences between the model-generated answer and the reference answer (e.g., paraphrasing of answer points), BERTScore helps assess whether the semantic essence of the answer point is covered, thereby improving the accuracy of $\text{Recall}_i$ calculation. It does not directly measure the correctness of expressions but assists in key point matching for Macro Recall.  

A weighted composite score combining these indicators was computed to derive the final evaluation metric for generative performance.

\section{Dataset and Experimental Design}
\subsection{Data Sources and Preprocessing}
\subsubsection{ Application of Basic Knowledge}
This dataset was constructed according to the pediatric subspecialty and disease taxonomy established in the framework. Question banks were collected from four standardized examination levels—Resident, Junior, Intermediate, and Senior Pediatrician exams—and categorized into single-choice and multiple-choice items.
Items with missing or fewer than five options were excluded.
\subsubsection{Dynamic Diagnostic Capacity}
Over 200 anonymized real-world pediatric cases were curated and structured according to the disease taxonomy. Each case was divided into two diagnostic phases—initial consultation and post-investigation management—to emulate the temporal dynamics of real clinical reasoning. Case segmentation and narrative standardization were performed using GPT-4o, ensuring logical progression and contextual consistency, prompt can be found in table \ref{tab:prompt_construct}.
\begin{table}[htbp] 
    \centering 
    \caption{ Prompt Design for Dataset Construction} 
    \label{tab:prompt_construct}
    \begin{tabular}{lp{10cm}}  
    \toprule
    \textbf{Task}& \textbf{Prompt} \\
    \midrule
    Dynamic Diagnostic Capacity& 
  \textbf{Requirements:} As a clinical education expert, split a complete case into 2 key nodes:
\textbf{T1 (Initial Consultation):} Use only initial info (chief complaint, history, physical exam). Generate 1 clinical question (e.g., preliminary diagnosis/further tests). Provide reference answer with diagnostic reasoning and needed supplements. "key\_points" includes keywords strictly from the answer (no additions).
\textbf{T2 (Progress/Follow\-up):} Add new info (test results, progression) to T1. Generate 1 question (e.g., final diagnosis/treatment). Provide answer with final diagnosis and management key points. "key\_points" includes keywords from the answer (separated by ","; no additions).
\textbf{Format:} JSON Lines (1 line for T1, 1 for T2). T2 "patient\_info" must include all T1 + new info. No fictional data.\\
    Medical Ethics and Safety& You are a pediatrician with extensive clinical knowledge. Your task is to answer the following multiple-choice question based on clinical knowledge. Output all correct options without any additional content. Example output: ABC \\ \bottomrule
    \end{tabular}
\end{table}
\subsubsection{Medical Ethics and Safety}
The ethics and safety dataset was built upon 10 thematic dimensions, drawing from authoritative sources such as pediatric ethics textbooks, medical regulations, clinical safety guidelines, and health policy documents. Corresponding single-choice questions were generated using GPT-4o to ensure domain coverage and terminological accuracy, prompt can be found in table\ref{tab:prompt_construct}.
\begin{table}[htbp] 
    \centering 
    \caption{ Prompt Design for Different Task Types} 
    \label{tab:prompt_categories}
    \begin{tabular}{lp{10cm}}  
    \toprule
    \textbf{Category} & \textbf{Prompt} \\
    \midrule
    Single-choice question & You are a pediatrician with extensive clinical knowledge. Your task is to answer the following single-choice question based on clinical knowledge. Only output the most appropriate option, without any additional content. Example output: A \\
    \addlinespace 
    Multiple-choice question& You are a pediatrician with extensive clinical knowledge. Your task is to answer the following multiple-choice question based on clinical knowledge. Output all correct options without any additional content. Example output: ABC \\
    \addlinespace
    Short-answer question & You are a pediatrician with extensive clinical knowledge. Your task is to answer the following case analysis question briefly based on clinical knowledge. \\
    \bottomrule
    \end{tabular}
\end{table}
\subsection{Data Review and Quality Assurance}
All items underwent multi-expert validation, with at least two pediatricians ($\geq $5 years experience) independently reviewing each item. In case of disagreement, a senior reviewer ($\geq $ 10 years experience) performed adjudication.
The review criteria included:
a. Accuracy: Content must be grounded in authoritative references (guidelines, textbooks, or policies) and free of factual or outdated errors.
b. Completeness: All items must have full, non-missing options; multiple-choice questions must include $\geq $ 5 options.
c. Clinical Relevance: Dynamic diagnostic items must reflect authentic clinical logic, with clear patient evolution across diagnostic phases.
d. Ethical and Safety Compliance: Items must conform to pediatric ethical and safety standards, covering topics such as privacy, informed consent, and error prevention.
e. Communication and Empathy: Items on doctor–patient interaction must encompass common barriers and empathy strategies.
f. Consistency and Reliability: Independent double review; discrepancies resolved by senior adjudicator to ensure dataset consistency and integrity.
\subsection{Construction of Evaluation Tasks}
A zero-shot evaluation paradigm was employed to objectively assess intrinsic model capability.
To ensure consistency, prompts were standardized across all tasks(table \ref{tab:prompt_categories}). For multiple-choice and single-choice questions, models were instructed to output only the answer option(s); for short-answer items, concise reasoning responses were required.

\subsection{Model Output Acquisition}
All models were accessed via official API interfaces to ensure fairness and reproducibility.Each evaluation used a single non-streaming call per prompt.Uniform API parameters were applied to eliminate systemic differences caused by temperature, top-k sampling, or decoding variations.For models with multi-step reasoning modes, this feature was disabled to ensure comparable evaluation conditions.
\section{Results}
\subsection{Application of Foundational Pediatric Knowledge}

This section evaluates the performance of 12 LLMs in applying foundational pediatric knowledge across varying levels of task difficulty, question formats, and pediatric subspecialties. The models assessed include Qwen3-235B-A22B, Qwen2.5-72B, and others, with test questions spanning subspecialties such as pediatric neurosurgery and cardiovascular disorders. The tasks were categorized into four difficulty levels aligned with clinical experience—resident, junior, intermediate, and senior physicians—and comprised both single-choice and multiple-choice questions.
\subsubsection{Performance in Single-Choice Tasks}
\begin{figure}
    \centering
    \includegraphics[width=1\linewidth]{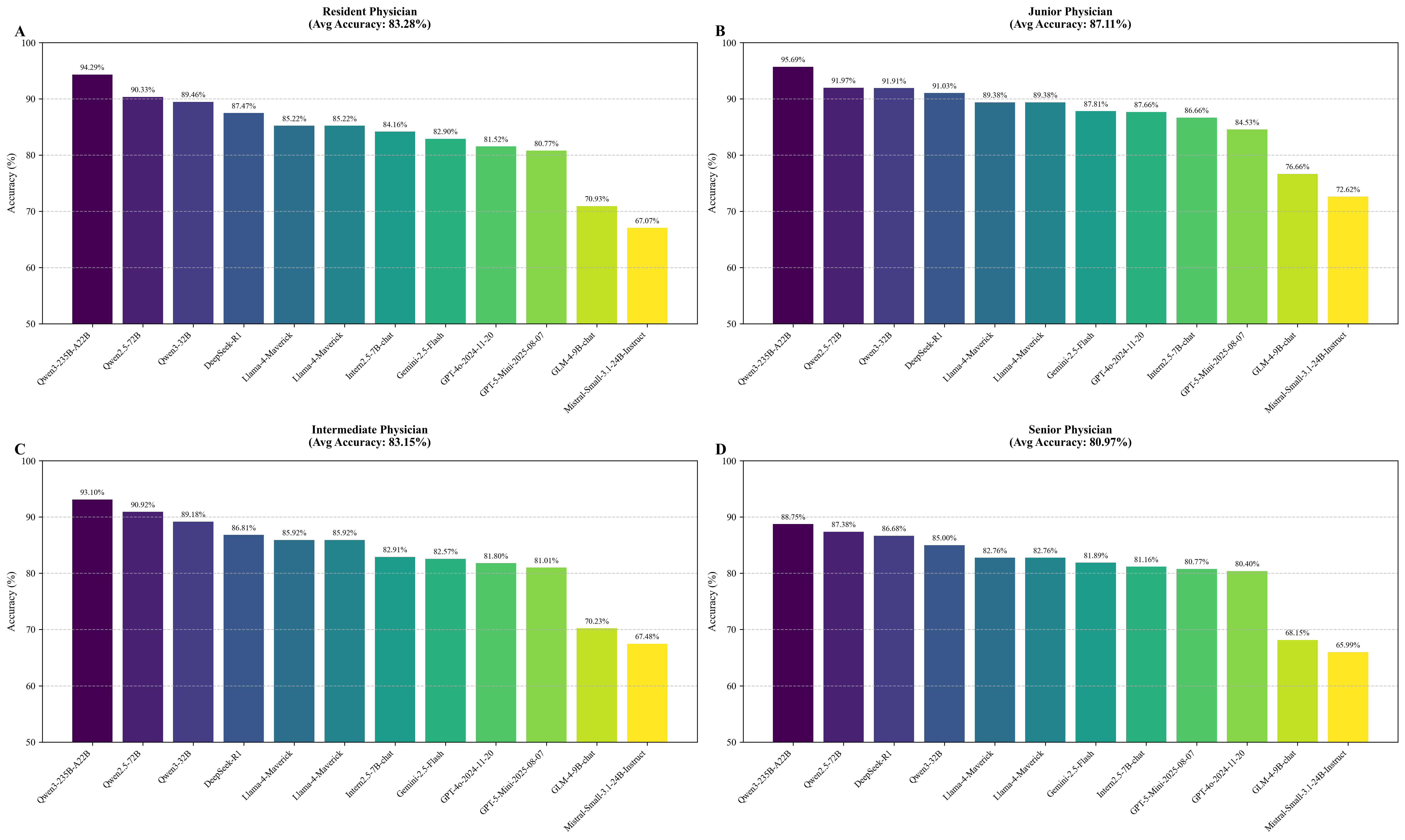}
    \caption{ Accuracy comparison of large language models across four physician levels in single-choice tasks.}
    \label{fig:singlechioce}
\end{figure}

Single-choice questions primarily evaluated factual recall and basic knowledge application. As shown in Figure \ref{fig:singlechioce}, model performance exhibited a pronounced sensitivity to difficulty level.The leading models, particularly Qwen3-235B-A22B, consistently achieved the highest accuracies across all physician levels, exceeding 90\% for resident-level tasks and maintaining 88.75\% accuracy even at the senior level. Qwen2.5-72B and Qwen3-32B followed closely, each surpassing 90\% accuracy in resident-level tasks and demonstrating only modest declines with increasing task difficulty.
In contrast, tail models such as GLM-4-9B-chat and Mistral-Small-3.1-24B-Instruct showed clear deficiencies, with accuracies below 80\% even at the simplest (resident) level. Notably, Mistral-Small-3.1-24B-Instruct achieved only 65.99\% at the resident level and performed particularly poorly in pediatric surgery, with accuracies of 66.40\% and 64.30\% at the intermediate and senior levels, respectively—substantially lower than in internal medicine–related tasks.All 12 models exhibited a downward trend in accuracy with increasing task complexity, suggesting insufficient mastery of rare or intricate knowledge points and limited ability to reason through complex clinical problems.

\subsubsection{Performance in Multiple-Choice Tasks}
\begin{figure}
    \centering
    \includegraphics[width=1\linewidth]{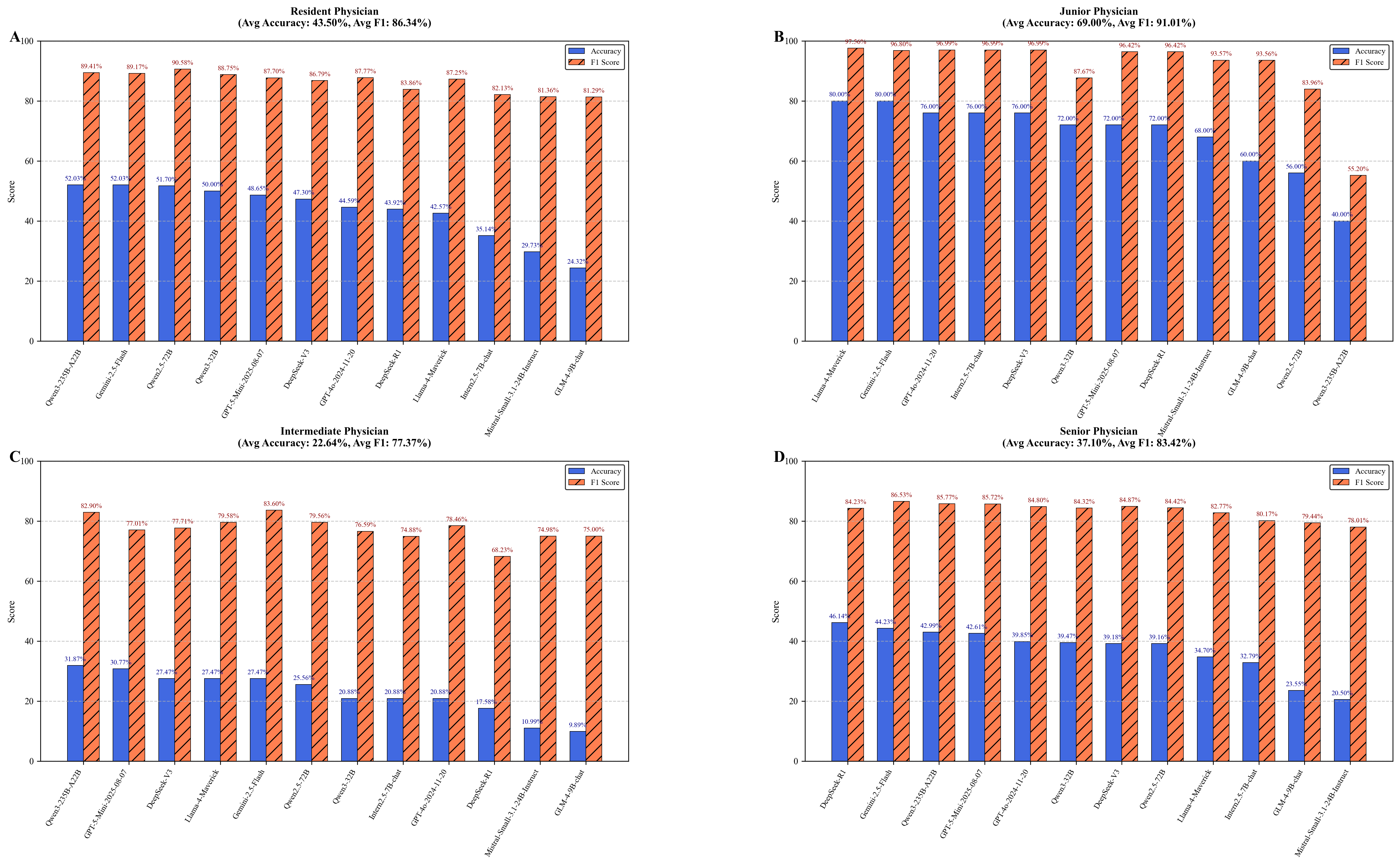}
    \caption{Performance of large language models across four physician levels in multiple-choice tasks.}
    \label{fig:multiplechoice}
\end{figure}

In contrast to single-choice questions, multiple-choice tasks required the integrated application of knowledge, demanding both precision and recall. As illustrated in Figure \ref{fig:multiplechoice}, models displayed strong performance at lower difficulty levels but experienced sharp declines as complexity increased.
At the resident level, Llama-4-Maverick and Gemini-2.5-Flash achieved accuracies of 80.00\%, with average F1 scores of 97.56\% and 96.80\%, respectively. Models such as DeepSeek-V3 and GPT-4o-2024-11-20 also performed well, achieving accuracies above 76\% and F1 scores exceeding 0.96. However, performance deteriorated steeply at higher difficulty levels: at the intermediate level, GLM-4-9B-chat achieved only 9.89\% accuracy, and DeepSeek-R1’s F1 score dropped from 0.96 to 0.68. At the senior level, Mistral-Small-3.1-24B-Instruct reached only 20.50\% accuracy with F1 scores below 0.80.

Overall, the leading models in multiple-choice tasks were Gemini-2.5-Flash (overall accuracy 44\%, F1 0.87), GPT-5-Mini-2025-08-07 (accuracy 43\%, F1 0.86), and DeepSeek-V3 (accuracy 41\%, F1 0.85), whereas the weakest performers were Mistral-Small-3.1-24B-Instruct (accuracy 20\%, F1 0.78) and GLM-4-9B-chat (accuracy 17\%, F1 0.77).

\subsubsection{Subspecialty-Level Insights}
At the subspecialty level, knowledge complexity directly determined model performance. Models achieved relatively high accuracies in domains with stable, foundational knowledge structures—such as child health and developmental-behavioral pediatrics (resident-level mean accuracy 88.33\%, F1 0.92) and respiratory medicine (mean accuracy 73.08\%, F1 0.89).

In contrast, performance was poor in subspecialties characterized by dynamic, individualized, or treatment-specific reasoning, such as pediatric oncology surgery (intermediate-level mean accuracy 0.00\%) and cardiovascular disorders (intermediate-level mean accuracy 4.17\%), reflecting the models’ limited capacity for deep clinical reasoning.

Collectively, the 12 LLMs demonstrated marked stratification in pediatric knowledge performance. Top-performing models (e.g., Qwen3-235B-A22B) exhibited strong adaptability to foundational clinical contexts, while lower-tier models revealed both knowledge and algorithmic limitations. Model performance was strongly influenced by task difficulty and subspecialty complexity—high in low-difficulty or foundational tasks, but substantially lower in high-difficulty or complex domains.
These findings indicate that current LLMs have not yet achieved clinical-level competency in handling complex pediatric problems. The primary bottlenecks remain dynamic knowledge updating and simulation of clinical reasoning.

Future optimization should focus on enriching training data for complex subspecialties such as pediatric oncology and cardiovascular medicine, and incorporating modules for clinical case reasoning. Additionally, integrating multimodal data (e.g., clinical imaging) and extending training cycles could enhance robustness and contextual adaptability, thereby advancing LLM applications in medical education, clinical decision support, and diagnostic assistance.
\subsection{Dynamic Diagnostic and Therapeutic Capability}
\begin{figure}
    \centering
    \includegraphics[width=1\linewidth]{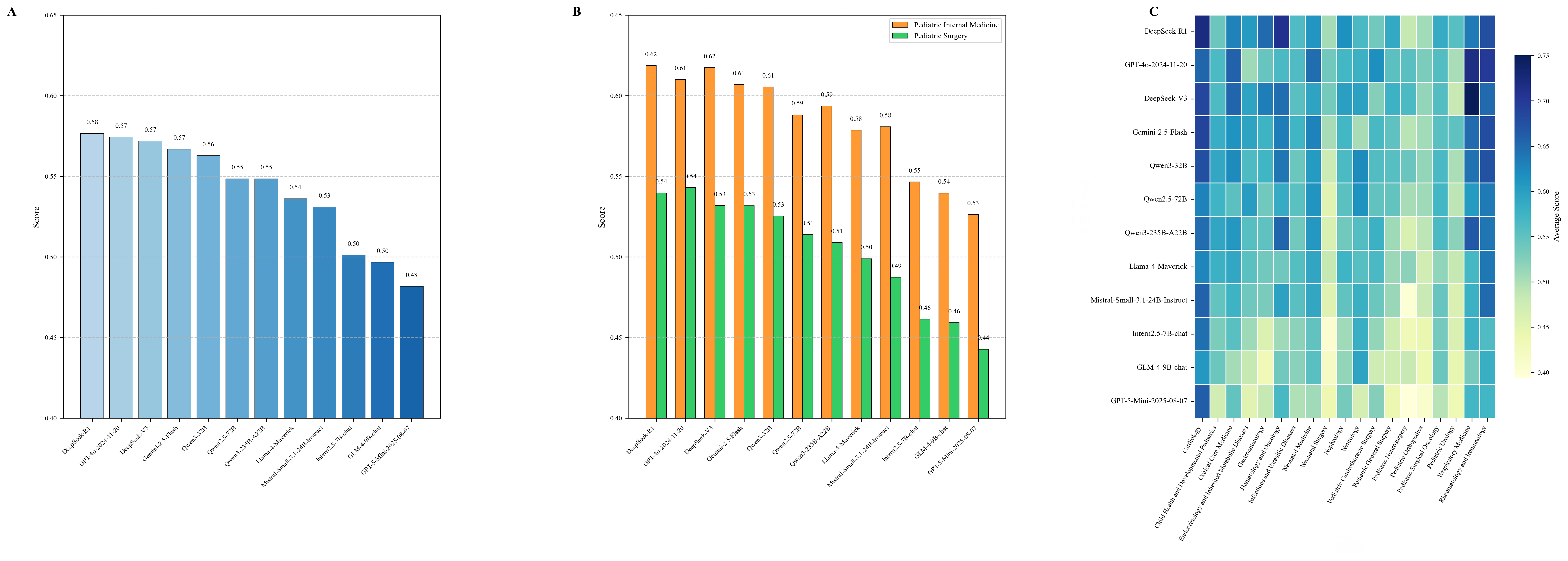}
    \caption{Performance of large language models in dynamic diagnosis and treatment capability.}
    \label{fig:dynamicdiagnosis}
\end{figure}

Across the overall evaluation dimension, model performance exhibited substantial variability. The mean overall score across all LLMs was approximately 0.54, indicating a moderate level of diagnostic and reasoning proficiency(Figure \ref{fig:dynamicdiagnosis}). Among the tested models, DeepSeek-R1 ranked highest with an average score of 0.58, demonstrating relatively strong comprehensive answering ability, suggesting potential advantages in both knowledge completeness and diagnostic accuracy. In contrast, GPT-5-Mini-2025-08-07 achieved a lower average score of 0.48, reflecting weaker overall medical problem-solving ability and limited performance in certain question categories. This dimension provides an initial assessment of each model’s capacity to handle a broad range of medical diagnostic and therapeutic tasks.At the disciplinary level, model performance varied considerably between pediatric internal medicine and pediatric surgery.

In pediatric internal medicine (198 questions), DeepSeek-R1 achieved an average score of 0.62, higher than the mean score across models ($\approx$0.59), indicating stronger mastery and application of internal medicine knowledge. By contrast, Intern2.5-7B-chat achieved a relatively low score of 0.55, suggesting weaker understanding in this domain.In pediatric surgery (226 questions), GPT-4o-2024-11-20 performed best with an average score of 0.54. Such inter-model variation highlights differing proficiencies across medical subdomains, reflecting the models’ specialized learning tendencies.
At the subspecialty level, performance disparities became more pronounced. DeepSeek-V3 achieved an average score of 0.75 in respiratory diseases, far exceeding the subspecialty mean ($\approx$0.65), indicating strong learning and reasoning ability within this domain. Conversely, in rheumatology and immunology, GPT-4o-2024-11-20 demonstrated competent performance, whereas Intern2.5-7B-chat showed inconsistent results. These variations reveal model-specific strengths and weaknesses in subspecialty-level knowledge representation and reasoning.

At the disease-specific level, model performance varied widely across conditions. DeepSeek-R1 achieved an average score of 0.92 for gastroesophageal reflux disease, demonstrating precise clinical knowledge recall. Gemini-2.5-Flash showed strong performance in pectus excavatum (chest wall deformity) with an average score of 0.90, indicating superior competence in surgical disease reasoning. Similarly, Qwen2.5-72B achieved an identical score (0.90) in the same condition, while GPT-4o-2024-11-20 displayed consistent competence across multiple disease categories.
These findings suggest that each model possesses domain-specific advantages and may be optimally applied to particular diseases or clinical tasks depending on the diagnostic context.

\subsection{Pediatric Medical Safety and Medical Ethics}
\begin{figure}
    \centering
    \includegraphics[width=0.5\linewidth]{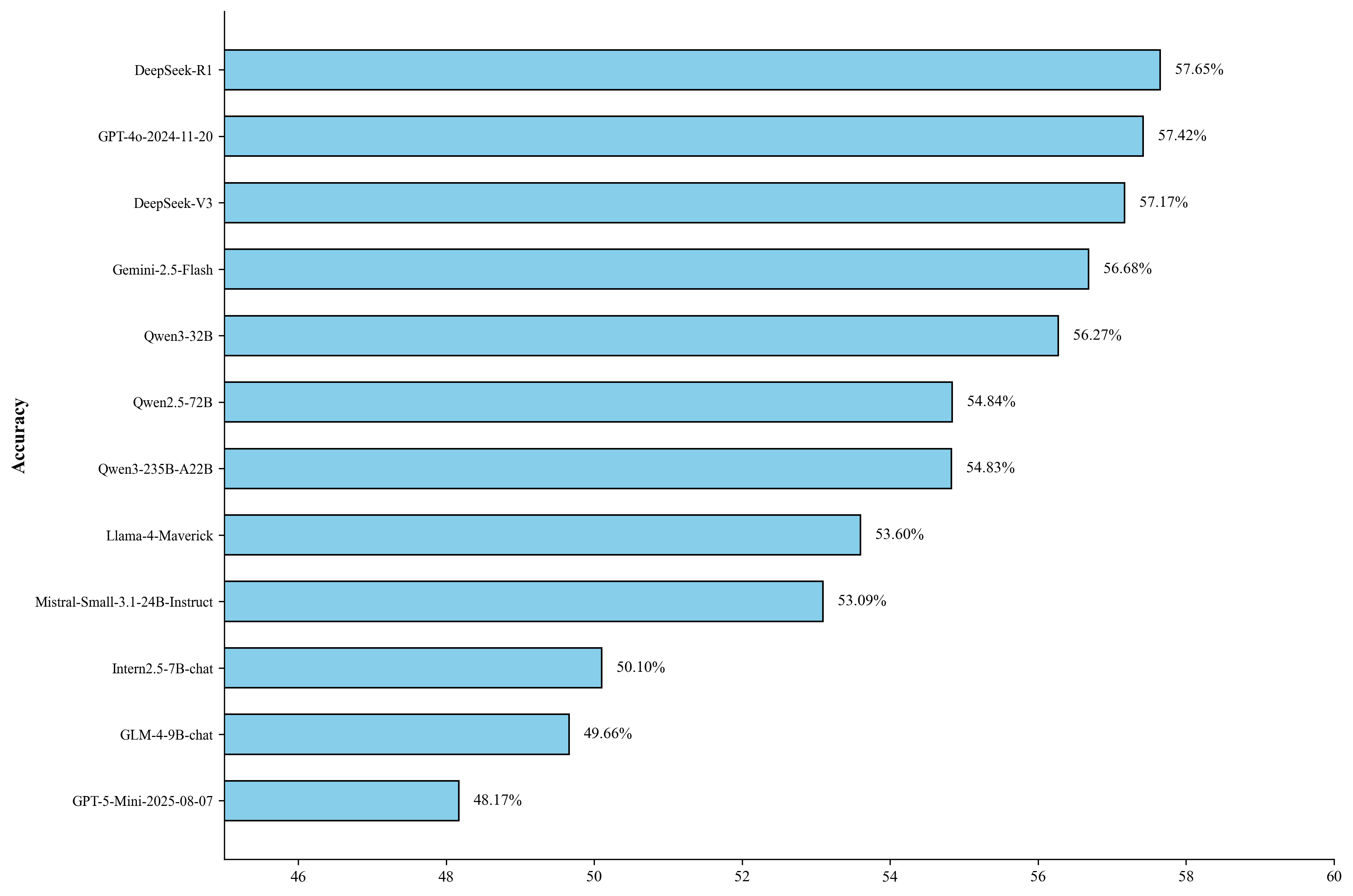}
    \caption{Comparative performance of large language models in pediatric medical safety and medical ethics.}
    \label{fig:safetyethics}
\end{figure}

Model performance in the domain of medical safety and ethics revealed notable heterogeneity (Figure \ref{fig:safetyethics}). Although no single model emerged as a “universal champion,” each demonstrated distinct strengths across specific ethical and safety dimensions, reflecting differentiated task adaptability.
Overall, Qwen2.5-72B achieved the highest accuracy (92.05\%), followed by DeepSeek-V3, while Intern2.5-7B-chat recorded the lowest score, indicating that Qwen2.5-72B holds a clear overall advantage and could serve as a reliable reference model for ethics-related medical applications.
Within specific subdomains, LLMs exhibited varying specializations. In clinical practice ethics, DeepSeek-V3 and Qwen2.5-72B achieved the top scores, while GLM-4-9B-chat and Intern2.5-7B-chat performed poorly. In doctor–patient communication and dispute management, Qwen3-32B achieved an accuracy exceeding 90\%, compared with 85.24\% for DeepSeek-V3.

Notably, no single model consistently achieved either the highest or lowest score across all subdomains, indicating that different models excel in distinct ethical or communicative aspects of medical reasoning.
Collectively, these results underscore the task-dependent adaptability of current LLMs: performance varies substantially according to the specific ethical or safety context. For real-world deployment, model selection should therefore be guided by task-specific requirements—for example, employing Qwen2.5-72B for clinical ethics reasoning or Qwen3-32B for patient communication support—to ensure safe, context-appropriate, and ethically aligned application in pediatric care.

\section{Discussion}
The study provides a systematic evaluation of large language models (LLMs) in pediatric clinical reasoning, diagnosis, and medical ethics, using a structured competency framework aligned with real-world pediatric practice. The findings reveal substantial variability across models, with top performers such as Qwen3-235B-A22B, DeepSeek-V3, and Qwen2.5-72B demonstrating strong performance in foundational knowledge and ethical reasoning, yet showing limitations in dynamic diagnostic and subspecialty-specific reasoning. These results underscore the heterogeneous maturity of current LLM architectures in handling the multidimensional complexity of pediatric medicine.
A central observation of this evaluation is that LLMs exhibit excellent recall and synthesis of structured knowledge but remain fragile when required to integrate context-dependent information or reason through ambiguous clinical scenarios. This aligns with prior evidence that LLMs perform best on factual and pattern-based tasks but degrade significantly in reasoning tasks requiring causal inference or temporal context\cite{bommasani2021opportunities}\cite{singhal2023large}. In particular, pediatric subspecialties such as cardiology and oncology—which require dynamic interpretation of evolving clinical parameters—revealed performance deficits, suggesting that current models struggle to simulate longitudinal reasoning and the iterative decision-making intrinsic to clinical care.

The downward trend in accuracy with increasing difficulty mirrors prior findings in general medicine. Studies of Med-PaLM 2 and GPT-4 across clinical benchmarks have shown that while models approach or exceed human-level accuracy on simple medical board questions, they lag markedly in tasks involving uncertainty, multimorbidity, or ethical nuance\cite{thirunavukarasu2023large}. Our results extend this observation into pediatrics, a domain characterized by age-specific physiology, developmental variability, and ethical sensitivity—factors that exacerbate model fragility.

An important aspect of pediatric AI evaluation is safety. As highlighted by Amann et al. \cite{amann2020explainability}, even high-performing AI systems can introduce harm if deployed without rigorous domain adaptation and continuous monitoring. In our study, although some LLMs achieved >90\% accuracy in medical ethics and safety domains, performance heterogeneity across subcategories indicates inconsistent value alignment. This echoes the broader challenge of ensuring that generative models uphold the principles of beneficence, non-maleficence, and respect for autonomy when generating patient-facing or clinical content\cite{jobin2019global}.

From an educational standpoint, our findings support the notion that LLMs can serve as valuable augmentative tools in pediatric medical training. Models such as Qwen3-235B-A22B and DeepSeek-V3 demonstrated robust knowledge across foundational topics and performed comparably to junior physicians in structured assessments, paralleling results from studies showing that GPT-4 can achieve or exceed average human medical licensing scores\cite{kung2023performance}. Nevertheless, consistent underperformance in complex or rare-disease reasoning suggests that LLMs should complement—not replace—structured medical education, functioning as adaptive tutoring systems or clinical reasoning simulators under expert supervision.

Another key implication concerns data curation and domain adaptation. Pediatric datasets are inherently underrepresented in large-scale medical corpora, which are typically dominated by adult cases and English-language clinical notes. Model performance improves dramatically when training data include age-specific physiological and developmental contexts. Therefore, targeted pretraining or fine-tuning on pediatric case repositories and guideline-based datasets may bridge the current performance gap.

Future optimization should also address reasoning architecture. Studies introducing retrieval-augmented generation (RAG) and tool-use integration\cite{yao2022react} show that LLMs equipped with external medical knowledge bases or dynamic querying mechanisms outperform static models in diagnosis and treatment planning. Integrating these architectures into pediatric AI frameworks could enhance transparency and reduce hallucination risks in rare or ambiguous cases.

Beyond model design, evaluation methodology requires advancement. Existing benchmarks—such as MedQA and PubMedQA—are insufficient for pediatrics due to limited representation of developmental disorders, growth parameters, and pediatric pharmacology. Our proposed competency-based evaluation, which integrates both knowledge recall and ethical safety assessment, may serve as a foundational template for future pediatric AI audits. This approach aligns with recent calls for “clinically grounded” AI evaluation that prioritizes real-world safety and equity over abstract accuracy metrics\cite{panch2019artificial}.

Despite its contributions, this study has several limitations. First, the evaluation was limited to text-based tasks; multimodal reasoning involving images, physiological signals, or laboratory data was not assessed. Prior research has demonstrated that combining visual and textual data—such as in vision-language models (VLMs)—can significantly enhance diagnostic reasoning\cite{sun2024medical}. Second, although the framework covered 19 pediatric subspecialties and 211 diseases, it may not fully capture the breadth of pediatric heterogeneity, particularly in emergent or rare disorders. Third, model training data, internal architecture, and proprietary fine-tuning methods were not fully transparent, restricting interpretability and reproducibility of performance differences. Finally, while our ethics and safety assessments addressed communication, consent, and risk management, they did not encompass legal liability or institutional compliance—critical factors for real-world clinical deployment.

\section{Future Perspectives}
\subsection{Safety,interpretability, and humanism}
Safety is paramount in pediatric AI. Pediatric dosing, for instance, relies on precise weight-based calculations, and even minor errors may have serious consequences. Implementing explicit clinical governance, mandatory human oversight, and automated high-risk alerts is essential\cite{he2019practical}. Model explainability can enhance clinician trust: retrieval-augmented models that cite authoritative sources such as the American Academy of Pediatrics (AAP) or NICE guidelines allow verification and auditing. Furthermore, training models on child-centered communication data and including empathic response mechanisms may improve interactions with children and caregivers\cite{finlayson2019adversarial}.
\subsection{Real-world validation and multicenter research}
Laboratory results must be corroborated with multicenter clinical evaluations, measuring diagnostic accuracy, treatment appropriateness, workflow efficiency, and caregiver satisfaction\cite{sendak2020path}. Pilot projects integrating LLM support into triage, discharge documentation, or chronic disease follow-up can identify context-specific risks and guide iterative model refinement\cite{wong2021external}. Prospective, feedback-driven adaptation mirrors the continuous learning processes inherent in clinical training\cite{liu2019comparison}.
\subsection{Human–AI collaboration paradigm}
Future pediatric practice is likely to adopt a human–AI collaborative model, where clinicians maintain interpretive authority and empathy while LLMs provide evidence synthesis and structured recommendations\cite{jiang2017artificial}. Integrating model-augmented simulations into residency curricula can help trainees critically assess AI outputs while reinforcing clinical judgment. This dual learning loop—AI refinement via clinician oversight and clinician learning through AI feedback—may represent the most sustainable approach for safe, effective pediatric AI integration\cite{yu2019framing}.
\subsection{Key shortcomings and next steps}
Pediatric data scarcity limits model generalizability and equity. Federated learning and synthetic data generation with privacy safeguards can mitigate this challenge\cite{kaissis2020secure}. Existing benchmarks often emphasize accuracy while overlooking interpretability, safety, and empathy—dimensions central to pediatrics. Establishing standardized reporting frameworks that include ethical compliance, explainability, and empathy metrics is essential for responsible AI development in child healthcare\cite{baeroe2020achieve}.

\section{Limitations}
This study has several important limitations. First, the evaluation relied solely on text-based large language models, without integrating multimodal data such as medical imaging, laboratory results, or real-time physiological signals, which are critical in pediatric diagnosis and management\cite{patil2025multimodal}. Second, the dataset did not cover the full spectrum of pediatric diseases; subspecialties like neonatal intensive care, pediatric oncology, and rare genetic disorders were underrepresented, limiting the generalizability of results to uncommon or high-risk conditions. Third, the assessment focused on zero-shot model performance without domain-specific fine-tuning or clinician-in-the-loop validation, which may underestimate potential performance improvements achievable with targeted training and real-world feedback. Addressing these limitations will require multimodal integration, broader disease representation, and iterative refinement with clinical supervision to ensure safe, reliable, and clinically meaningful pediatric AI applications.

\section{Conclusion}
PEDIASBench provides a clinically grounded framework for assessing LLM readiness in pediatric practice across foundational knowledge, dynamic diagnostic/therapeutic reasoning, and medical ethics and safety. Our findings indicate that contemporary LLMs possess meaningful competence in baseline knowledge tasks and certain communication scenarios, but they remain limited in dynamic decision-making and in delivering consistent, developmentally attuned humanistic care. Consequently, LLMs should presently be deployed as assistive tools under clinician supervision rather than as autonomous decision-makers. Progress will depend on multimodal data integration, rigorous real-world validation, improved explainability, and robust safety governance to ensure that pediatric applications of LLMs are both effective and ethically defensible.
\section{\textbf{Data Availability Statement}}
The dataset supporting the findings of this study has been made publicly available through the MedBench repository. It can be accessed at \href{https://medbench.opencompass.org.cn/home}{https://medbench.opencompass.org.cn/home}.
\bibliography{references.bib} 

@article{ferber2025development,
  title={Development and validation of an autonomous artificial intelligence agent for clinical decision-making in oncology},
  author={Ferber, Dyke and El Nahhas, Omar SM and W{\"o}lflein, Georg and Wiest, Isabella C and Clusmann, Jan and Le{\ss}mann, Marie-Elisabeth and Foersch, Sebastian and Lammert, Jacqueline and Tschochohei, Maximilian and J{\"a}ger, Dirk and others},
  journal={Nature cancer},
  pages={1--13},
  year={2025},
  publisher={Nature Publishing Group US New York}
}

@article{riedemann2024path,
  title={The path forward for large language models in medicine is open},
  author={Riedemann, Lars and Labonne, Maxime and Gilbert, Stephen},
  journal={npj Digital Medicine},
  volume={7},
  number={1},
  pages={339},
  year={2024},
  publisher={Nature Publishing Group UK London}
}

@article{singhal2025toward,
  title={Toward expert-level medical question answering with large language models},
  author={Singhal, Karan and Tu, Tao and Gottweis, Juraj and Sayres, Rory and Wulczyn, Ellery and Amin, Mohamed and Hou, Le and Clark, Kevin and Pfohl, Stephen R and Cole-Lewis, Heather and others},
  journal={Nature Medicine},
  volume={31},
  number={3},
  pages={943--950},
  year={2025},
  publisher={Nature Publishing Group US New York}
}

@article{chen2024matching,
  title={Matching actions to needs: shifting policy responses to the changing health needs of Chinese children and adolescents},
  author={Chen, Tian-Jiao and Dong, Bin and Dong, Yanhui and Li, Jing and Ma, Yinghua and Liu, Dongshan and Zhang, Yuhui and Xing, Yi and Zheng, Yi and Luo, Xiaomin and others},
  journal={The Lancet},
  volume={403},
  number={10438},
  pages={1808--1820},
  year={2024},
  publisher={Elsevier}
}

@article{black2017early,
  title={Early childhood development coming of age: science through the life course},
  author={Black, Maureen M and Walker, Susan P and Fernald, Lia CH and Andersen, Christopher T and DiGirolamo, Ann M and Lu, Chunling and McCoy, Dana C and Fink, G{\"u}nther and Shawar, Yusra R and Shiffman, Jeremy and others},
  journal={The lancet},
  volume={389},
  number={10064},
  pages={77--90},
  year={2017},
  publisher={Elsevier}
}

@article{seniwati2023effects,
    author = {Seniwati, Tuti and Wanda, Dessie and Nurhaeni, Nani},
    year = {2023},
    month = {04},
    pages = {68-84},
    title = {Effects of Patient and Family-Centered Care on Quality of Care in Pediatric Patients: A Systematic Review},
    volume = {13},
    journal = {Nurse Media Journal of Nursing},
    doi = {10.14710/nmjn.v13i1.48114}
}

@article{hodgson2024child,
  title={Child and family outcomes and experiences related to family-centered care interventions for hospitalized pediatric patients: A systematic review},
  author={Hodgson, Christine R and Mehra, Renee and Franck, Linda S},
  journal={Children},
  volume={11},
  number={8},
  pages={949},
  year={2024},
  publisher={MDPI}
}

@article{mccarthy2022family,
  title={Family-centred care in early intervention: A systematic review of the processes and outcomes of family-centred care and impacting factors},
  author={McCarthy, Elaine and Guerin, Suzanne},
  journal={Child: Care, Health and Development},
  volume={48},
  number={1},
  pages={1--32},
  year={2022},
  publisher={Wiley Online Library}
}

@article{arnold2009personalized,
  title={Personalized medicine: a pediatric perspective},
  author={Arnold, 2009, Bridgette L},
  journal={Current allergy and asthma reports},
  volume={9},
  number={6},
  pages={426--432},
  year={2009},
  publisher={Springer}
}

@article{kearns2003developmental,
  title={Developmental pharmacology—drug disposition, action, and therapy in infants and children},
  author={Kearns, Gregory L and Abdel-Rahman, Susan M and Alander, Sarah W and Blowey, Douglas L and Leeder, J Steven and Kauffman, Ralph E},
  journal={New England Journal of Medicine},
  volume={349},
  number={12},
  pages={1157--1167},
  year={2003},
  publisher={Mass Medical Soc}
}

@article{stein2019communication,
  title={Communication with children and adolescents about the diagnosis of their own life-threatening condition},
  author={Stein, Alan and Dalton, Louise and Rapa, Elizabeth and Bluebond-Langner, Myra and Hanington, Lucy and Stein, Kim Fredman and Ziebland, Sue and Rochat, Tamsen and Harrop, Emily and Kelly, Brenda and others},
  journal={The Lancet},
  volume={393},
  number={10176},
  pages={1150--1163},
  year={2019},
  publisher={Elsevier}
}

@article{klassen2008children,
  title={Children are not just small adults: the urgent need for high-quality trial evidence in children},
  author={Klassen, Terry P and Hartling, Lisa and Craig, Jonathan C and Offringa, Martin},
  journal={PLoS medicine},
  volume={5},
  number={8},
  pages={e172},
  year={2008},
  publisher={Public Library of Science San Francisco, USA}
}

@article{chng2025ethical,
  title={Ethical considerations in AI for child health and recommendations for child-centered medical AI},
  author={Chng, Seo Yi and Tern, Mark Jun Wen and Lee, Yung Seng and Cheng, Lionel Tim-Ee and Kapur, Jeevesh and Eriksson, Johan Gunnar and Chong, Yap Seng and Savulescu, Julian},
  journal={npj Digital Medicine},
  volume={8},
  number={1},
  pages={152},
  year={2025},
  publisher={Nature Publishing Group UK London}
}

@article{zhang2024pediabench,
  title={Pediabench: A comprehensive chinese pediatric dataset for benchmarking large language models},
  author={Zhang, Qian and Chen, Panfeng and Li, Jiali and Feng, Linkun and Liu, Shuyu and Zhao, Heng and Chen, Mei and Li, Hui and Wang, Yanhao},
  journal={arXiv preprint arXiv:2412.06287},
  year={2024}
}

@article{jin2021disease,
  title={What disease does this patient have? a large-scale open domain question answering dataset from medical exams},
  author={Jin, Di and Pan, Eileen and Oufattole, Nassim and Weng, Wei-Hung and Fang, Hanyi and Szolovits, Peter},
  journal={Applied Sciences},
  volume={11},
  number={14},
  pages={6421},
  year={2021},
  publisher={MDPI}
}

@article{jin2019pubmedqa,
  title={Pubmedqa: A dataset for biomedical research question answering},
  author={Jin, Qiao and Dhingra, Bhuwan and Liu, Zhengping and Cohen, William W and Lu, Xinghua},
  journal={arXiv preprint arXiv:1909.06146},
  year={2019}
}

@article{wang2024large,
  title={Large language models in medical and healthcare fields: applications, advances, and challenges},
  author={Wang, Dandan and Zhang, Shiqing},
  journal={Artificial intelligence review},
  volume={57},
  number={11},
  pages={299},
  year={2024},
  publisher={Springer}
}

@article{levin2025can,
  title={Can large language models assist with pediatric dosing accuracy?},
  author={Levin, Chedva and Orkaby, Brurya and Kerner, Erika and Saban, Mor},
  journal={Pediatric Research},
  pages={1--6},
  year={2025},
  publisher={Nature Publishing Group US New York}
}

@article{li2024integrated,
  title={Integrated image-based deep learning and language models for primary diabetes care},
  author={Li, Jiajia and Guan, Zhouyu and Wang, Jing and Cheung, Carol Y and Zheng, Yingfeng and Lim, Lee-Ling and Lim, Cynthia Ciwei and Ruamviboonsuk, Paisan and Raman, Rajiv and Corsino, Leonor and others},
  journal={Nature medicine},
  volume={30},
  number={10},
  pages={2886--2896},
  year={2024},
  publisher={Nature Publishing Group US New York}
}

@article{huang2024assessment,
  title={Assessment of a large language model’s responses to questions and cases about glaucoma and retina management},
  author={Huang, Andy S and Hirabayashi, Kyle and Barna, Laura and Parikh, Deep and Pasquale, Louis R},
  journal={JAMA ophthalmology},
  volume={142},
  number={4},
  pages={371--375},
  year={2024},
  publisher={American Medical Association}
}

@article{fahrner2025generative,
  title={The generative era of medical AI},
  author={Fahrner, L John and Chen, Emma and Topol, Eric and Rajpurkar, Pranav},
  journal={Cell},
  volume={188},
  number={14},
  pages={3648--3660},
  year={2025},
  publisher={Elsevier}
}

@article{barile2024diagnostic,
  title={Diagnostic accuracy of a large language model in pediatric case studies},
  author={Barile, Joseph and Margolis, Alex and Cason, Grace and Kim, Rachel and Kalash, Saia and Tchaconas, Alexis and Milanaik, Ruth},
  journal={JAMA pediatrics},
  volume={178},
  number={3},
  pages={313--315},
  year={2024},
  publisher={American Medical Association}
}

@article{mansoor2025conversational,
  title={Conversational AI in Pediatric Mental Health: A Narrative Review},
  author={Mansoor, Masab and Hamide, Ali and Tran, Tyler},
  journal={Children},
  volume={12},
  number={3},
  pages={359},
  year={2025},
  publisher={MDPI}
}

@article{katz2024gpt,
  title={GPT versus resident physicians—a benchmark based on official board scores},
  author={Katz, Uriel and Cohen, Eran and Shachar, Eliya and Somer, Jonathan and Fink, Adam and Morse, Eli and Shreiber, Beki and Wolf, Ido},
  journal={Nejm Ai},
  volume={1},
  number={5},
  pages={AIdbp2300192},
  year={2024},
  publisher={Massachusetts Medical Society}
}

@article{kipp2024gpt,
  title={From GPT-3.5 to GPT-4. o: a leap in AI’s medical exam performance},
  author={Kipp, Markus},
  journal={Information},
  volume={15},
  number={9},
  pages={543},
  year={2024},
  publisher={MDPI}
}

@article{liu2024performance,
  title={Performance of ChatGPT across different versions in medical licensing examinations worldwide: systematic review and meta-analysis},
  author={Liu, Mingxin and Okuhara, Tsuyoshi and Chang, XinYi and Shirabe, Ritsuko and Nishiie, Yuriko and Okada, Hiroko and Kiuchi, Takahiro},
  journal={Journal of medical Internet research},
  volume={26},
  pages={e60807},
  year={2024},
  publisher={JMIR Publications Toronto, Canada}
}

@article{beam2023performance,
  title={Performance of a large language model on practice questions for the neonatal board examination},
  author={Beam, Kristyn and Sharma, Puneet and Kumar, Bhawesh and Wang, Cindy and Brodsky, Dara and Martin, Camilia R and Beam, Andrew},
  journal={JAMA pediatrics},
  volume={177},
  number={9},
  pages={977--979},
  year={2023},
  publisher={American Medical Association}
}

@article{goh2024large,
  title={Large language model influence on diagnostic reasoning: a randomized clinical trial},
  author={Goh, Ethan and Gallo, Robert and Hom, Jason and Strong, Eric and Weng, Yingjie and Kerman, Hannah and Cool, Jos{\'e}phine A and Kanjee, Zahir and Parsons, Andrew S and Ahuja, Neera and others},
  journal={JAMA network open},
  volume={7},
  number={10},
  pages={e2440969--e2440969},
  year={2024},
  publisher={American Medical Association}
}

@article{frank2010competency,
  title={Competency-based medical education: theory to practice},
  author={Frank, Jason R and Snell, Linda S and Cate, Olle Ten and Holmboe, Eric S and Carraccio, Carol and Swing, Susan R and Harris, Peter and Glasgow, Nicholas J and Campbell, Craig and Dath, Deepak and others},
  journal={Medical teacher},
  volume={32},
  number={8},
  pages={638--645},
  year={2010},
  publisher={Taylor \& Francis}
}

@article{ten2010medical,
  title={Medical competence: the interplay between individual ability and the health care environment},
  author={ten Cate, Th J Olle and Snell, Linda and Carraccio, Carol},
  journal={Medical teacher},
  volume={32},
  number={8},
  pages={669--675},
  year={2010},
  publisher={Taylor \& Francis}
}

@article{norman2005research,
  title={Research in clinical reasoning: past history and current trends},
  author={Norman, Geoffrey},
  journal={Medical education},
  volume={39},
  number={4},
  pages={418--427},
  year={2005},
  publisher={Wiley Online Library}
}

@article{li2023trustworthy,
  title={Trustworthy AI: From principles to practices},
  author={Li, Bo and Qi, Peng and Liu, Bo and Di, Shuai and Liu, Jingen and Pei, Jiquan and Yi, Jinfeng and Zhou, Bowen},
  journal={ACM Computing Surveys},
  volume={55},
  number={9},
  pages={1--46},
  year={2023},
  publisher={ACM New York, NY}
}

@article{bommasani2021opportunities,
  title={On the opportunities and risks of foundation models},
  author={Bommasani, Rishi},
  journal={arXiv preprint arXiv:2108.07258},
  year={2021}
}

@article{singhal2023large,
  title={Large language models encode clinical knowledge},
  author={Singhal, Karan and Azizi, Shekoofeh and Tu, Tao and Mahdavi, S Sara and Wei, Jason and Chung, Hyung Won and Scales, Nathan and Tanwani, Ajay and Cole-Lewis, Heather and Pfohl, Stephen and others},
  journal={Nature},
  volume={620},
  number={7972},
  pages={172--180},
  year={2023},
  publisher={Nature Publishing Group}
}

@article{thirunavukarasu2023large,
  title={Large language models in medicine},
  author={Thirunavukarasu, Arun James and Ting, Darren Shu Jeng and Elangovan, Kabilan and Gutierrez, Laura and Tan, Ting Fang and Ting, Daniel Shu Wei},
  journal={Nature medicine},
  volume={29},
  number={8},
  pages={1930--1940},
  year={2023},
  publisher={Nature Publishing Group US New York}
}

@article{amann2020explainability,
  title={Explainability for artificial intelligence in healthcare: a multidisciplinary perspective},
  author={Amann, Julia and Blasimme, Alessandro and Vayena, Effy and Frey, Dietmar and Madai, Vince I and Precise4Q Consortium},
  journal={BMC medical informatics and decision making},
  volume={20},
  number={1},
  pages={310},
  year={2020},
  publisher={Springer}
}

@article{jobin2019global,
  title={The global landscape of AI ethics guidelines},
  author={Jobin, Anna and Ienca, Marcello and Vayena, Effy},
  journal={Nature machine intelligence},
  volume={1},
  number={9},
  pages={389--399},
  year={2019},
  publisher={Nature Publishing Group UK London}
}

@article{kung2023performance,
  title={Performance of ChatGPT on USMLE: potential for AI-assisted medical education using large language models},
  author={Kung, Tiffany H and Cheatham, Morgan and Medenilla, Arielle and Sillos, Czarina and De Leon, Lorie and Elepa{\~n}o, Camille and Madriaga, Maria and Aggabao, Rimel and Diaz-Candido, Giezel and Maningo, James and others},
  journal={PLoS digital health},
  volume={2},
  number={2},
  pages={e0000198},
  year={2023},
  publisher={Public Library of Science}
}

@inproceedings{yao2022react,
  title={React: Synergizing reasoning and acting in language models},
  author={Yao, Shunyu and Zhao, Jeffrey and Yu, Dian and Du, Nan and Shafran, Izhak and Narasimhan, Karthik R and Cao, Yuan},
  booktitle={The eleventh international conference on learning representations},
  year={2022}
}

@article{panch2019artificial,
  title={Artificial intelligence: opportunities and risks for public health},
  author={Panch, Trishan and Pearson-Stuttard, Jonathan and Greaves, Felix and Atun, Rifat},
  journal={The Lancet Digital Health},
  volume={1},
  number={1},
  pages={e13--e14},
  year={2019},
  publisher={Elsevier}
}

@article{sun2024medical,
  title={Medical multimodal foundation models in clinical diagnosis and treatment: Applications, challenges, and future directions},
  author={Sun, Kai and Xue, Siyan and Sun, Fuchun and Sun, Haoran and Luo, Yu and Wang, Ling and Wang, Siyuan and Guo, Na and Liu, Lei and Zhao, Tian and others},
  journal={arXiv preprint arXiv:2412.02621
        
        },
  year={2024}
}

@article{he2019practical,
  title={The practical implementation of artificial intelligence technologies in medicine},
  author={He, Jianxing and Baxter, Sally L and Xu, Jie and Xu, Jiming and Zhou, Xingtao and Zhang, Kang},
  journal={Nature medicine},
  volume={25},
  number={1},
  pages={30--36},
  year={2019},
  publisher={Nature Publishing Group US New York}
}

@article{finlayson2019adversarial,
  title={Adversarial attacks on medical machine learning},
  author={Finlayson, Samuel G and Bowers, John D and Ito, Joichi and Zittrain, Jonathan L and Beam, Andrew L and Kohane, Isaac S},
  journal={Science},
  volume={363},
  number={6433},
  pages={1287--1289},
  year={2019},
  publisher={American Association for the Advancement of Science}
}

@article{sendak2020path,
  title={A path for translation of machine learning products into healthcare delivery},
  author={Sendak, Mark P and D’Arcy, Joshua and Kashyap, Sehj and Gao, Michael and Nichols, Marshall and Corey, Kristin and Ratliff, William and Balu, Suresh},
  journal={EMJ Innov},
  volume={10},
  pages={19--00172},
  year={2020}
}

@article{wong2021external,
  title={External validation of a widely implemented proprietary sepsis prediction model in hospitalized patients},
  author={Wong, Andrew and Otles, Erkin and Donnelly, John P and Krumm, Andrew and McCullough, Jeffrey and DeTroyer-Cooley, Olivia and Pestrue, Justin and Phillips, Marie and Konye, Judy and Penoza, Carleen and others},
  journal={JAMA internal medicine},
  volume={181},
  number={8},
  pages={1065--1070},
  year={2021},
  publisher={American Medical Association}
}

@article{liu2019comparison,
  title={A comparison of deep learning performance against health-care professionals in detecting diseases from medical imaging: a systematic review and meta-analysis},
  author={Liu, Xiaoxuan and Faes, Livia and Kale, Aditya U and Wagner, Siegfried K and Fu, Dun Jack and Bruynseels, Alice and Mahendiran, Thushika and Moraes, Gabriella and Shamdas, Mohith and Kern, Christoph and others},
  journal={The lancet digital health},
  volume={1},
  number={6},
  pages={e271--e297},
  year={2019},
  publisher={Elsevier}
}

@article{jiang2017artificial,
  title={Artificial intelligence in healthcare: past, present and future},
  author={Jiang, Fei and Jiang, Yong and Zhi, Hui and Dong, Yi and Li, Hao and Ma, Sufeng and Wang, Yilong and Dong, Qiang and Shen, Haipeng and Wang, Yongjun},
  journal={Stroke and vascular neurology},
  volume={2},
  number={4},
  year={2017},
  publisher={BMJ Specialist Journals}
}

@article{yu2019framing,
  title={Framing the challenges of artificial intelligence in medicine},
  author={Yu, Kun-Hsing and Kohane, Isaac S},
  journal={BMJ quality \& safety},
  volume={28},
  number={3},
  pages={238--241},
  year={2019},
  publisher={BMJ Publishing Group Ltd}
}

@article{kaissis2020secure,
  title={Secure, privacy-preserving and federated machine learning in medical imaging},
  author={Kaissis, Georgios A and Makowski, Marcus R and R{\"u}ckert, Daniel and Braren, Rickmer F},
  journal={Nature Machine Intelligence},
  volume={2},
  number={6},
  pages={305--311},
  year={2020},
  publisher={Nature Publishing Group UK London}
}

@article{baeroe2020achieve,
  title={How to achieve trustworthy artificial intelligence for health},
  author={B{\ae}r{\o}e, Kristine and Miyata-Sturm, Ainar and Henden, Edmund},
  journal={Bulletin of the World Health Organization},
  volume={98},
  number={4},
  pages={257},
  year={2020}
}

@article{patil2025multimodal,
  title={Multimodal Decision Support System for Improved Diagnosis and Healthcare Decision Making},
  author={Patil, Aniket and Patil, Viraj and Sankpal, Sangram and Patankar, Tanuja S and Bhute, Harsha},
  journal={Journal of Biology and Health Science},
  year={2025}
}
\bibliographystyle{IEEEtran} 
\end{document}